# Optimization Algorithm-Based Approach for Modelling Large Deflection of Cantilever Beam Subjected to Tip Load

Fei Gao, Gaoyu Liu, Wei-Hsin Liao[1]

Department of Mechanical and Automation Engineering, The Chinese University of Hong Kong, Shatin, NT, Hong Kong, China

**Abstract**

Beam mechanism and beam theory have attracted substantial attention from researchers, as they have been widely used in many fields such as compliant mechanisms and soft robots. The modeling of beam mechanisms becomes complicated due to the geometric nonlinearity that is proved to be significant with large deflection. A new method, called optimization algorithm-based approach (OABA), is proposed to predict the large deflection of cantilever beams, in which an optimization algorithm is exploited to find the locus of the beam tip. With the derived locus of the beam tip, the deflection curve of the cantilever beam can be calculated. The optimization algorithm in this paper is embodied in a particle swarm optimization (PSO) algorithm. Experimental results show that the proposed method can precisely predict the deflection of the uniform and non-uniform cantilever beams. The maximum error is limited to 4.35% when the normalized maximum transverse deflection reaches 0.75. Given that, the proposed OABA would be a universal approach to model the large deflection of cantilever beams subjected to tip loads.

## 1. Introduction

Beam mechanism and beam theory have been widely used in many engineering fields such as compliant mechanisms [1-7], flexible hinges [8-15], soft robots [16-21], piezoelectric beam-based energy harvesters [22-25], and leaf springs [26-27]. Compliant mechanisms exploit the deflection of the beams to transfer or transform the smooth and precise force and motion. Owing to utilizing fewer moveable parts, compliant mechanisms have several merits over the traditional rigid-body mechanisms such as low backlash and low wear. These advantages mentioned have attracted considerable attention from researchers and industrials, and most of them devoted to promoting the study and use of compliant mechanisms over the last decades. Leaf springs can be more compact and lightweight, compared with coil springs, for providing the same stiffness and energy storage capability, so as to enable moveable robots to provide the required elasticity at the same time meet the size and weight limitations.

To predict/characterize the deformation of beam mechanisms, vast researchers committed themselves to devise and study the beam's governing load-deformation relations. Linear beam theory has been developed and widely used to derive the small deformation of cantilever beams subjected to tip loads. However, once the deformation becomes large, nonlinearity will be raised in the governing equations due to arc-length conservation and geometric nonlinearity. The linear beam theory is not valid in this case. To handle that, researchers proposed alternative methods to model the large deflection of a cantilever beam. These currently available methods can be generally categorized as elliptical integration solution, finite element method, beam constraint model, energy-minimization-based solution, and pseudo-rigid body model [28]. The elliptic integration solution is considered as the most accurate one, which is used as a standard solution to evaluate other methods. It is noteworthy that the axial elongation and shear of the beam are not considered in the elliptic integration solutions. Lyon and Howell [29] developed the elliptic integration solution with no inflection point for a fixed-fixed beam, then Kimball and Tsai [30] added an inflection point to the elliptic integration solution. Zhang and Chen [31] established a comprehensive elliptic integration solution with the consideration of the number of inflection points. With this method, the deflection of the cantilever beam can be precisely estimated regardless of the load conditions and deflection modes.

Beam constraint model (BCM) [32] is a closed-form and parametric mathematical model, which is used to capture the constraint characteristics of a cantilever beam in terms of the stiffness and errors motion. Besides, in this model, the load conditions, initial and boundary conditions, and beam shapes can be taken into account. Further, the nonlinearity associated with load equilibrium is also considered here. The accuracy of this method can be guaranteed over a range of load and displacement. In particular, the maximum transverse deflection is limited to ±0.1. Then Awtar and Sen [33] developed a nonlinear strain energy formulation of the BCM based on principles of virtual work to derive the nonlinear load-displacement relations for complex flexure mechanisms. Here the internal interaction force and the load equilibrium for each constitute beam can be ignored thus significantly simplifying the calculation. To increase the range of deflection, Chen et al. [34-35] developed a chained beam constant model (CBCM). The flexible beam was divided into a few elements, and each element is characterized by BCM.

---
[1] Corresponding author.
Email address: whliao@cuhk.edu.hk

Howell and Midha [36] proposed to approximate the beam's deflection with a pseudo-rigid body model (PRB 1R). Specifically, the cantilever beam is modeled as two rigid links and an intermediate torsion spring. One of the rigid links is fixed to the ground, and the other one is connected to the first one by a pin joint. The length of each rigid link and the stiffness of the torsion spring are optimized to reduce the tip locus errors between the derived results and the numerical elliptic integral data when the cantilever beam is subject to a pure end force. To improve the accuracy and expand the load conditions, Kimball and Tsai [37] developed PRB 2R that consists of three rigid links for modeling the deflection of a cantilever beam with an inflection point. Yu et al. [38] also developed a PRB 2R to improve the accuracy of PRB 1R. However, the parameters in the PRB 2R model are dependent on the loading conditions. Su [39] established a PRB 3R model with high accuracy for a large range of deflection. Besides, the parameters in this model are optimized to obtain high accuracy regardless of the load conditions [40-41]. To accurately predict the deflection of a cantilever beam with an inflection point, Yu et al. [42-43] developed a PRB 5R, in which six rigid links are joint at five pin joints. Four of the pin joints are accompanied by torsion springs, and the rest one is a free hinge. Unlike the aforementioned PRB 1R, Verotti [44] developed a new one-DoF rigid body model, in which the stiffness coefficient was evaluated based on strain energy, and the occurrence of an inflection point was also considered here.

The aforementioned methods exploited governing equations or approximated mathematic models to derive the large deflection of a cantilever beam. Note that these methods are highly dependent on the load conditions, boundary conditions, or assumptions for simplification. It is hard to guarantee the high accuracy of these methods when the cantilever beam is subjected to a large range of load conditions. Besides, the modification of these methods for different load conditions and the calculation/implementation are complicated and tedious. In the last decades, artificial intelligence (AI) encountered great progress. Researchers have proposed to use AI such as neural networks and optimization algorithms to solve engineering problems to which traditional approaches are ineffective or infeasible. The details of the systems such as the boundary conditions or load conditions in the above models can be ignored, and the solutions can be obtained through performing sets of iterations. Shahabi and Kuo [45] devised an artificial neural network (ANN) to solve the inverse kinematics solutions of the dual-backbone continuum robot, and the forward kinematics solutions derived from PRBM were used to train the ANN. Mohamad et al. [46] employed a PSO to identify the parameters in the dynamic model of a flexible beam structure. Saffar et al. [47] developed an ANN to model the experimental data of an aluminum cantilever beam. With this trained model, the natural frequencies of the system can be predicted.

The objective of this paper is to propose a new approach to model the large deflection of a cantilever beam when subjected to a combined tip load. Unlike the existing methods, this new method, called OABA, utilized an optimization algorithm to find the locus of the beam tip through performing iteration processes. The optimization algorithm is embodied in a PSO algorithm in this study. Then with that and based on the Euler-Bernoulli (E-B) beam theory, the deflection curve of the cantilever beam can be derived. The initial state of the cantilever beam is being straight, and the cross-section of the beam can be uniform and non-uniform along the length. The work presented here is organized in the following sequence. Section 2 introduces the working principles of the PSO algorithm, and then a PSO algorithm is utilized to find the locus of the beam tip. Section 3 describes the experimental validation of the presented approach. Section 4 discusses the extension of the presented method. For example, the presented OABA may be employed to derive the deflection of a cantilever beam with initial curvature. Finally, conclusions are made in terms of the presented study.

## 2. Particle swarm optimization-based approach to model large deflection of cantilever beam

### 2.1 Particle swarm optimization

Swarm intelligence was proposed based on the collective behaviour of decentralized and self-organized systems, which consist of a population of intelligent agents interacting with each other and the environment. Each agent searches the environment with a certain degree random, while the interaction/communication between each agent lead to "intelligence"-collective behaviour. Some well-known examples can clearly show the working principles of swarm intelligent systems, for example, ant colonies, bird flocking, and fish schooling. PSO is one kind of swarm intelligent algorithm, proposed by Kennedy and Eberhart [48]. PSO is a heuristic global optimization method, which performs searching processes via a swarm of particles that update their own positions and velocities based on individual particles' own searching experience and searching experience of the swarm. In the last decades, PSO was widely used in many engineering fields [49], such as electrical and electronic engineering [50-51], automatic control [52-53], and mechanical engineering [40, 54-55]. In our previous work, we have exploited the PSO algorithm to optimize the dimension of an MR damper thus decreasing the energy consumption at the same time providing the required damping force [56].

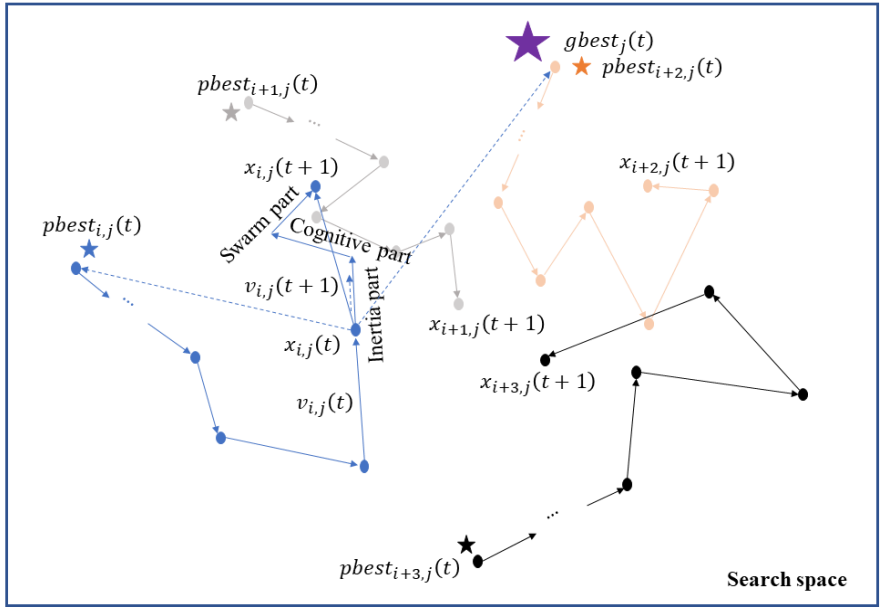

**Fig. 1 Schematic diagram of the update of the position and velocity of each particle in the swarm.**

PSO algorithm is a population-based search algorithm in which each particle wanders around the search space at an adaptable speed that is dynamically adjusted based on each particle's own search experience and the search experience of other particles. Each particle has two characteristics including velocity and position. The following formulations show the update of the velocity and position of each particle [55].

$$v_{i,j}(t+1) = wv_{i,j}(t) + c_1 r_1 \left(pbest_{i,j}(t) - x_{i,j}(t)\right) + c_2 r_2 \left(gbest_j(t) - x_{i,j}(t)\right) \quad (1)$$

$$x_{i,j}(t+1) = x_{i,j}(t) + v_{i,j}(t+1) \quad (2)$$

with $i = 1, 2, \cdots, n$ and $j = 1, 2, \cdots, m$.

where $w$ denotes the inertia weight used to balance the global search and local search. $c_1$ and $c_2$ are constant parameters, named, "acceleration coefficient". $r_1$ and $r_2$ are random variables with range $[0, 1]$. $pbest_{i,j}(t)$ refers to the best position of the $i^{th}$ particle in terms of the objective function value or fitness, and $gbest_j(t)$ is the best position out of all the swarm. The PSO algorithm consists of $n$ particles, and each particle can seek for the optimal solution in $m$-dimension parameter space. The velocity of the particle in $t + 1$ time step, expressed in equation (1), can be divided into three parts, as shown in **Fig. 1**. The first part is the momentum/inertia component that provides each particle with a trend to keep its previous velocity in the last time step. The second part refers to the cognitive/memory part that is related to the particle's own best position during all previous iterations. The final part is called the social/swarm part that drives the particle to get close to the best position of the swarm. The parameters shown in equation (1) have a great effect on the performance of the PSO algorithm, which should be carefully selected in practical use. In addition, the swarm size and stopping criteria also affect the performance in the implementation of the PSO algorithm.

The inertia weight $w$ is used to regulate the impact of the previous velocity on the current velocity, also is employed to control the trade-off between local search and global search. In this study, $w$ is not constant, but can be adaptively adjusted according to the particle's own fitness and the average and minimum fitness of the swarm.

$$w(t+1) = \begin{cases} w_{min} + (w_{max} - w_{min}) \frac{\left(fit(x_i(t)) - fit_{min}(t)\right)}{\left(fit_{ave}(t) - fit_{min}(t)\right)} & f_i(t) \leq f_{ave}(t) \\ w_{max} & f_i(t) > f_{ave}(t) \end{cases} \quad (3)$$

$$fit_{min}(t) = \min_{i=1,\cdots,n} \left[fit(x_i(t))\right] \quad (4)$$

$$fit_{ave}(t) = \frac{\sum_{i=1}^{n} fit(x_i(t))}{n} \tag{5}$$

where $w_{min}$ and $w_{max}$ refer to the minimum inertia weight and the maximum inertia weight, respectively. $fit(x_i(t))$ denotes the value of the fitness function of the $i^{th}$ particle.

For the implementation of the PSO algorithm, firstly the velocity and position of each particle will be initialized with random values. Then based on the values of the fitness function of each individual particle and the swarm, the velocity and position are updated in each time step, as illustrated in equations (1)-(2). The PSO algorithm will be terminated while the maximum iteration number or the required fitness is reached. Otherwise, the PSO algorithm will keep on performing iteration processes until the termination condition is fulfilled.

## 2.2 Modelling large deflection of cantilever beam

Most of the existing methods proposed to directly derive the deflection of cantilever beams based on some complex beam theory and specified boundary conditions [28-35]. Another widely used method called the PRB model approximates the flexible cantilever beam by several rigid links that are connected by pin joints and accompanied by torsion springs [36-44]. It is noteworthy that to obtain high accuracy, the parameters in the PRB model such as the length of each link and stiffness of the torsion springs are selected by optimization. However, in some special cases, this approach may not ensure acceptable accuracy. Unlike the existing methods, Here, in this study, we use a PSO algorithm to find the planar coordinates of the tip of a cantilever beam that subjects to tip loads. Then with the derived coordinates of the tip and based on the E-B beam theory, the deflection curve of the cantilever beam can be obtained. The schematic diagram of the optimization algorithm-based approach is presented in **Fig. 2**.

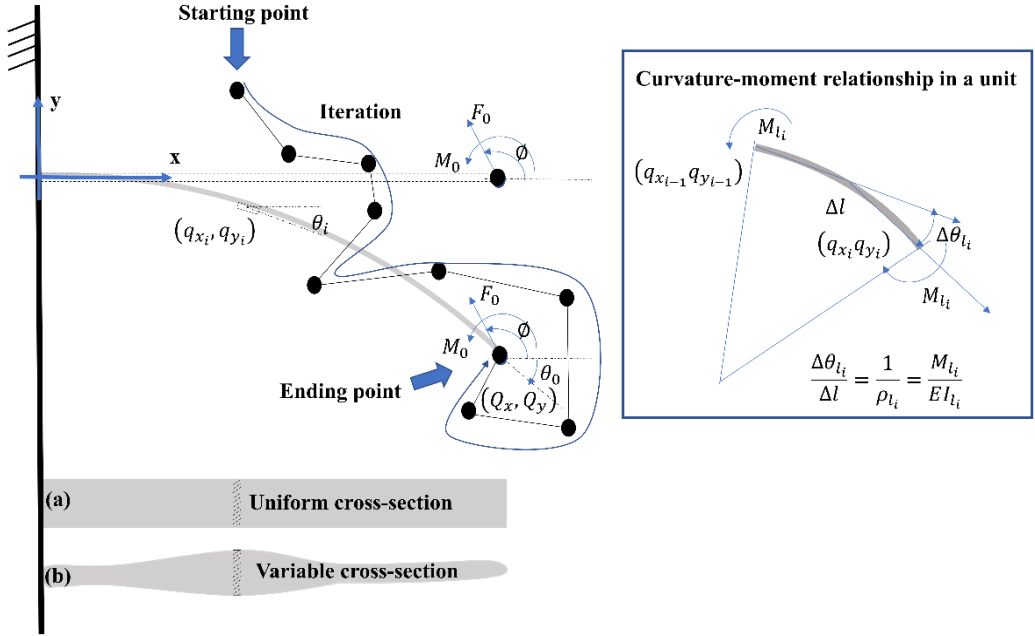

Fig. 2 Derivation of the locus of the tip when a cantilever beam is subjected to a combined tip load.

The cantilever beam is divided into $nl$ units, and the curvature-moment relationship in each unit is shown in **Fig. 2**. Here, the cantilever beam should be a long and slender beam, and only the bending deformation induced by the moment is considered in each unit. The extension and compression of each unit are ignored in this study. In addition, the cantilever beam has no initial curvature. The curvature of each unit is proportional to the bending moment, according to the classical Euler-Bernoulli theory.

$$\frac{\Delta\theta_{l_i}}{\Delta l} = \frac{1}{\rho_{l_i}} = \frac{M_{l_i}}{EI_{l_i}}, \quad i = 1,2,\cdots,nl \tag{6}$$

$$M_{l_i} = F_0 sin(\emptyset)(Q_x - q_{x_{i-1}}) - F_0 cos(\emptyset)(Q_y - q_{y_{i-1}}) + M_0 \tag{7}$$

where $\Delta l$ is the length of each unit. $\Delta\theta_{l_i}$ represents the variation of the slope in the $i^{th}$ unit. $\rho_{l_i}$ is the curvature radius of the $i^{th}$ unit. $E$ and $I_{l_i}$ refer to the Young's modulus and the moment of inertia of the $i^{th}$ unit, respectively. $M_{l_i}$ is the moment acting on the $i^{th}$ unit. $F_0$ and $\emptyset$ denote the tip force and the inclination angle of the tip force, respectively. $M_0$ refers to the tip moment. $(Q_x, Q_y)$ is the coordinates of the beam tip. $\theta_0$ represents the slope of the beam tip. The slope at the segment point $i+1^{th}$ is given by

$$\theta_{l_{i+1}} = \theta_{l_i} + \Delta\theta_{l_i} \tag{8}$$

As the stretching/compressing of the beam is ignored, the length of each unit is constant. The coordinates $(q_{x_{i+1}}, q_{y_{i+1}})$ of the unit point $i+1^{th}$ can be calculated

$$q_{x_{i+1}} = q_{x_i} + \Delta l \cos\theta_{l_i} \qquad (9)$$

$$q_{y_{i+1}} = q_{y_i} + \Delta l \sin\theta_{l_i} \qquad (10)$$

The unit model shown in equations (6)-(10) are linear, and the curvature of the unit is linearly proportional to the tip loads. The cantilever beam suffers from elastic deformation, and no plastic deformation is induced here. However, as the beam arc-length conservation will result in nonlinearity in the beam's geometry, the transverse deflection, longitudinal deflection, and displacement of the tip are nonlinearly proportional to the tip loads. The nonlinearity can be presented in the simulation and testing results, as presented in Section 3. In this study, each particle in the PSO algorithm consists of three elements including $Q_x$, $Q_y$, and $\theta_0$.

$$x_i(t) = [Q_{x-i}(t) \quad Q_{y-i}(t) \quad \theta_{0-i}(t)] \qquad (11)$$

Here, the fitness function in this PSO algorithm is given by

$$fit(x_i(t)) = \frac{\sqrt{(Q_x(t)-q_{x_{nl}}(t))^2 + (Q_y(t)-q_{y_{nl}}(t))^2}}{l} + \frac{|\theta_0(t)-\theta_{l_{nl}}(t)|}{2\pi} \qquad (12)$$

where $(q_{x_{nl}}(t), q_{y_{nl}}(t))$ are the derived coordinates of the beam tip. $(Q_x(t), Q_y(t))$ are generated and automatically updated by the PSO algorithm in each iteration. Substituting $(Q_x(t), Q_y(t))$ into equations (6)-(10), the $(q_{x_{nl}}(t), q_{y_{nl}}(t))$ can be calculated through performing *nl* iteration processes. The error between $(Q_x(t), Q_y(t))$ and $(q_{x_{nl}}(t), q_{y_{nl}}(t))$ is normalized with respect to the length of the cantilever beam *l*. Also, the error between $\theta_0(t)$ and $\theta_{l_{nl}}(t)$ is normalized by the parameter search range, $2\pi$. The particle's own best position $pbest_i(t)$, and the best position of the swarm $gbest(t)$ are

$$pbest_i(t) = arg \min_{k=1,\cdots,t} [fit(x_i(k))] \qquad (13)$$

$$gbest(t) = arg \min_{\substack{k=1,\cdots,t \\ i=1,\cdots,n}} [fit(x_i(k))] \qquad (14)$$

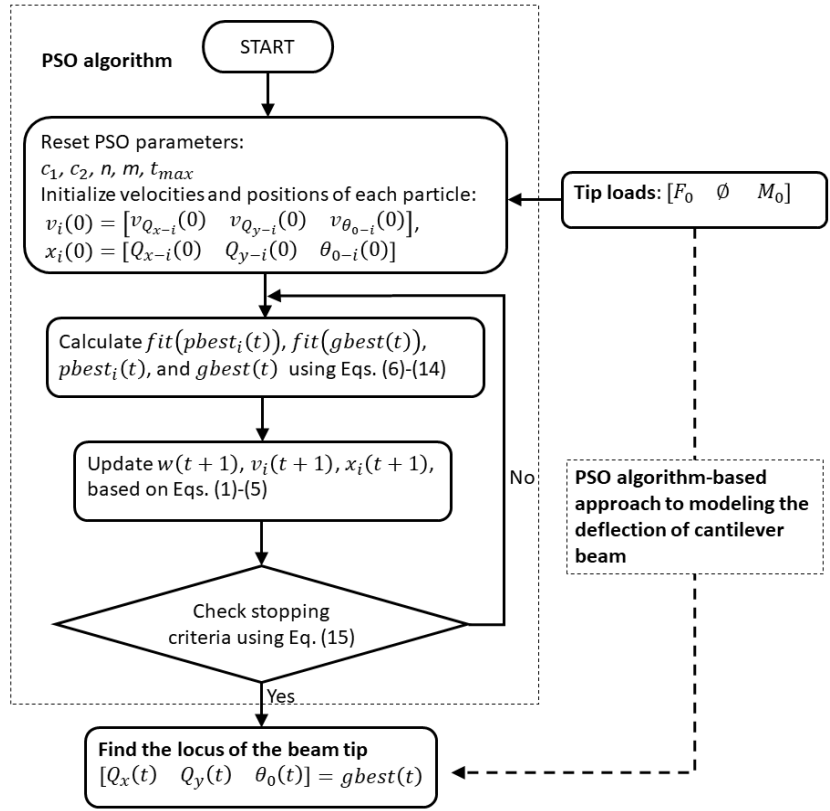

**Fig. 3 PSO algorithm-based approach for modelling the deflection of a cantilever beam.**

The stopping criteria used in this study is given by

$$fit(gbest(t)) \leq e_{th}, or\ t = t_{max} \qquad (15)$$

where $e_{th}$ is the acceptable error, and $t_{max}$ refers to the maximum iteration number. When the above condition is fulfilled, the iteration processes in the PSO algorithm will end.

With a given combined tip load $[F_0 \quad \emptyset \quad M_0]$, the derivation of the locus of the beam tip can be illustrated as follows:

1. Reset the values of the general parameters used in the PSO algorithm, $c_1$, $c_2$, n, m. Initialize each particle in the swarm with random velocities $v_i(0) = [v_{Q_{x-i}}(0) \quad v_{Q_{y-i}}(0) \quad v_{\theta_{0-i}}(0)]$ and random positions $x_i(0) = [Q_{x-i}(0) \quad Q_{y-i}(0) \quad \theta_{0-i}(0)]$. The velocities and positions should be located in their search space, respectively, as shown in Fig.1.
2. Substituting $x_i(0) = [Q_{x-i}(0) \quad Q_{y-i}(0) \quad \theta_{0-i}(0)]$ into equations (6)-(12), the fitness value of each particle can be calculated. Then taking those into equations (13)-(14), the individual best fitness value $fit(pbest_i(t))$ and global best fitness value $fit(gbest(t))$ and the corresponding position $pbest_i(t)$ and $gbest(t)$ can be derived.
3. With these derived parameters including $pbest_i(t)$ and $gbest(t)$, the inertia weight $w(t+1)$, velocities $v_i(t+1)$, and positions of each particle $x_i(t+1)$ can be updated in each iteration, according to equations (1)-(5).
4. Repeat steps 2 and 3 until the stopping criteria in equation (15) is achieved.
5. When the iteration is terminated, the coordinates and the slop of the beam tip are obtained, $[Q_x(t) \quad Q_y(t) \quad \theta_0(t)] = gbest(t)$. The derivation flow chart of the coordinates and the slop of the beam tip is presented in **Fig. 3**.

## 3. Experimental validation of the optimization algorithm-based approach

### 3.1 Deflection of uniform beam

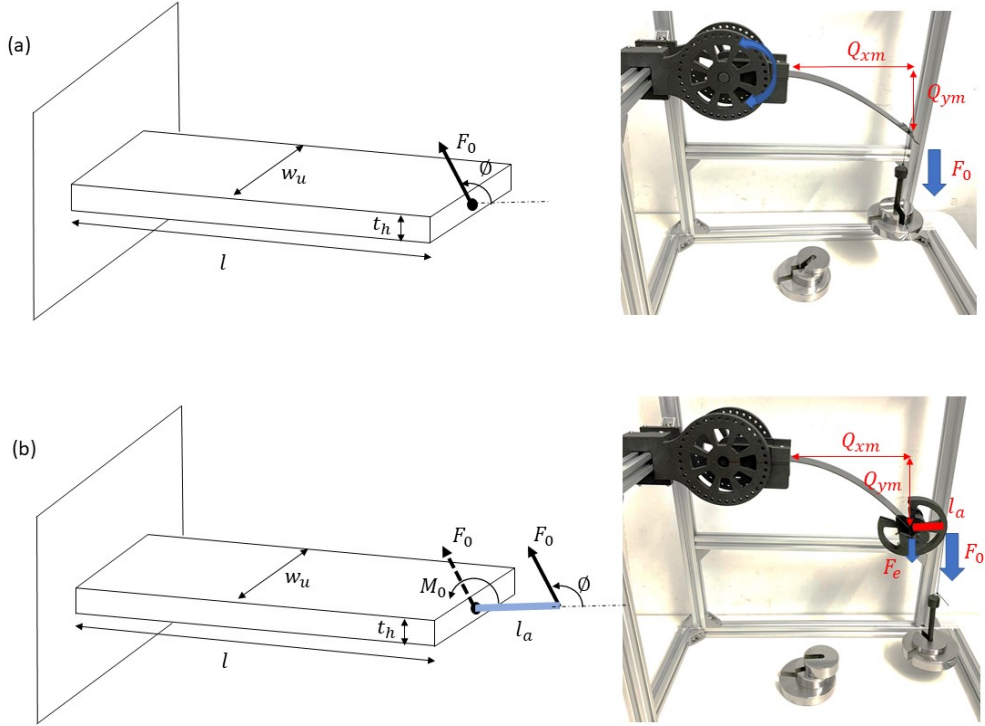

**Fig. 4 Deflection measurements of a uniform cantilever beam. (a) Subjected to a pure force load, (b) Subjected to a combined force and moment load.**

To validate the proposed method, we develop a platform, as shown in **Fig. 4**, for testing the deflection of a uniform beam. Then by comparing the tested results with the estimated results derived from the optimization algorithm-based approach, the accuracy can be calculated. Two cases including subjected to a pure force load and subjected to a combined force and moment load are investigated, as shown in **Fig.4**. The normalized error between the simulation and the testing results is given by

$$e_{norm} = \frac{error}{l} = \frac{\sqrt{(Q_x-Q_{xm})^2+(Q_y-Q_{ym})^2}}{l} \quad (16)$$

where $(Q_{xm}, Q_{ym})$ refers to the measured coordinates of the beam tip. During all testing, the maximum deformation of the cantilever beam will not exceed the maximum allowable elastic deformation. When the tip load is removed, the cantilever beam can restore to its initial shape. When the cantilever beam experienced a pure force load ($F_0 = 1.034\ N$), the PSO algorithm can efficiently find the locus of the beam tip after performing 26 iterations. The errors in each iteration and the derived locus of the beam tip are presented in **Fig. 5(a)** and **(b)**, respectively. Besides, substituting $(Q_x, Q_y)$ into equations (6)-(10), the deflection curves of the cantilever beam under different loads can be derived and presented in **Fig. 5(b)**. The values of the parameters used in this study are listed in **Table I**.

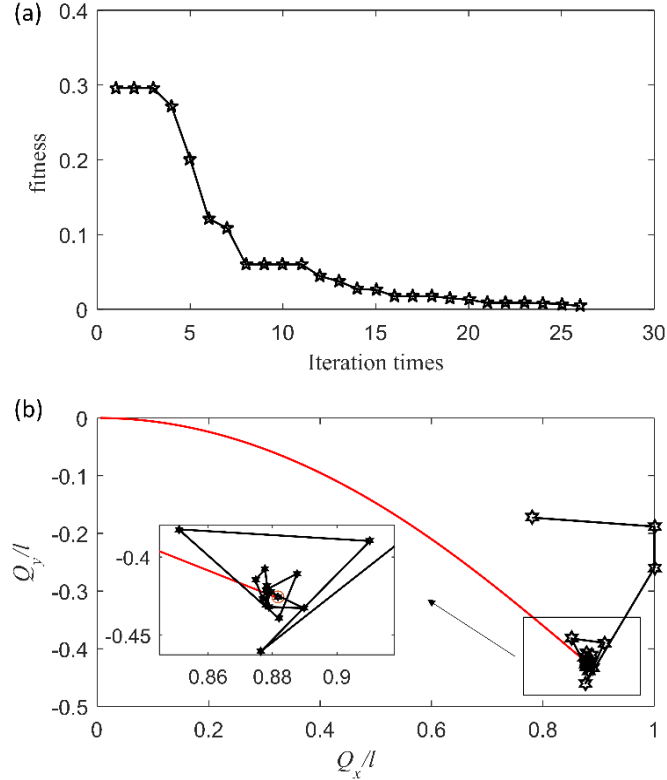

**Fig. 5** Derivation of the coordinates of beam tip based on the PSO algorithm. (a) Values of the fitness function in each iteration, (b) $gbest(t)$ in each iteration and the defection curve corresponding to the final $gbest$. ($F_0 = 6.958\ N, \emptyset = -\frac{5\pi}{6}, M_0 = 0\ Nm$)

**Table I**
Values of the parameters used in this study

| Parameters | Values |
| --- | --- |
| Length of the cantilever beam $l$ | 180 mm |
| Beam width (uniform beam) $w_u$ | 25 mm |
| Beam width (non-uniform beam) $w_{nu}$ | [22.6 26.9 30.0 22.8 22.8 26.8 28.1 20.8 23.7 28.1 29.6] mm |
| Beam thickness $t_h$ | 1.15 mm |
| Young's modulus $E$ | 45.36 GPa |
| Number of units $nl$ | 200 |
| Number of particles in the swarm $n$ | 100 |
| Acceleration coefficient $c_1$ | 0.2 |
| Acceleration coefficient $c_2$ | 0.2 |
| The maximum inertia weight $w_{max}$ | 0.8 |
| The minimum inertia weight $w_{min}$ | 0.6 |
| The maximum iteration number $t_{max}$ | 50 |
| Threshold for the fitness function $e_{th}$ | 0.005 |
| $Q_x$ search space | [-180 180] mm |
| $Q_y$ search space | [-180 180] mm |
| $\theta_0$ search space | [-π π] |

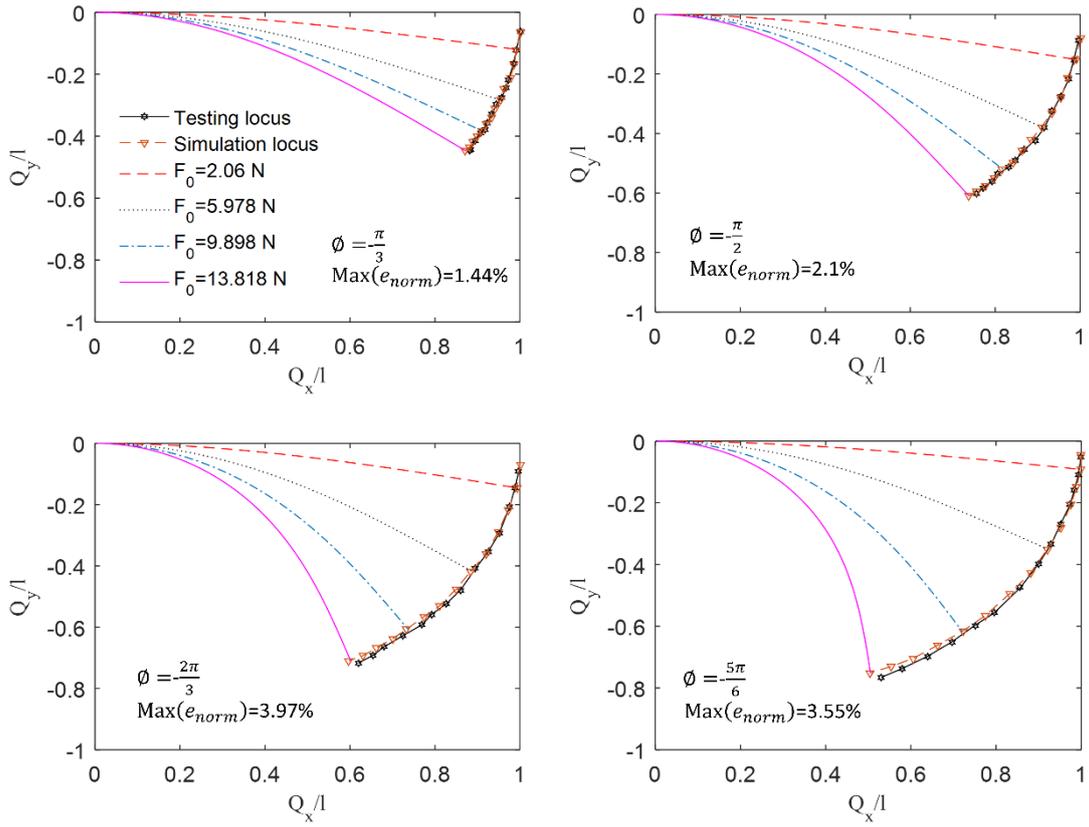

**Fig. 6 Comparison of the tip locus of the simulation and testing results. (Uniform cantilever beam subjected to a pure force load)**

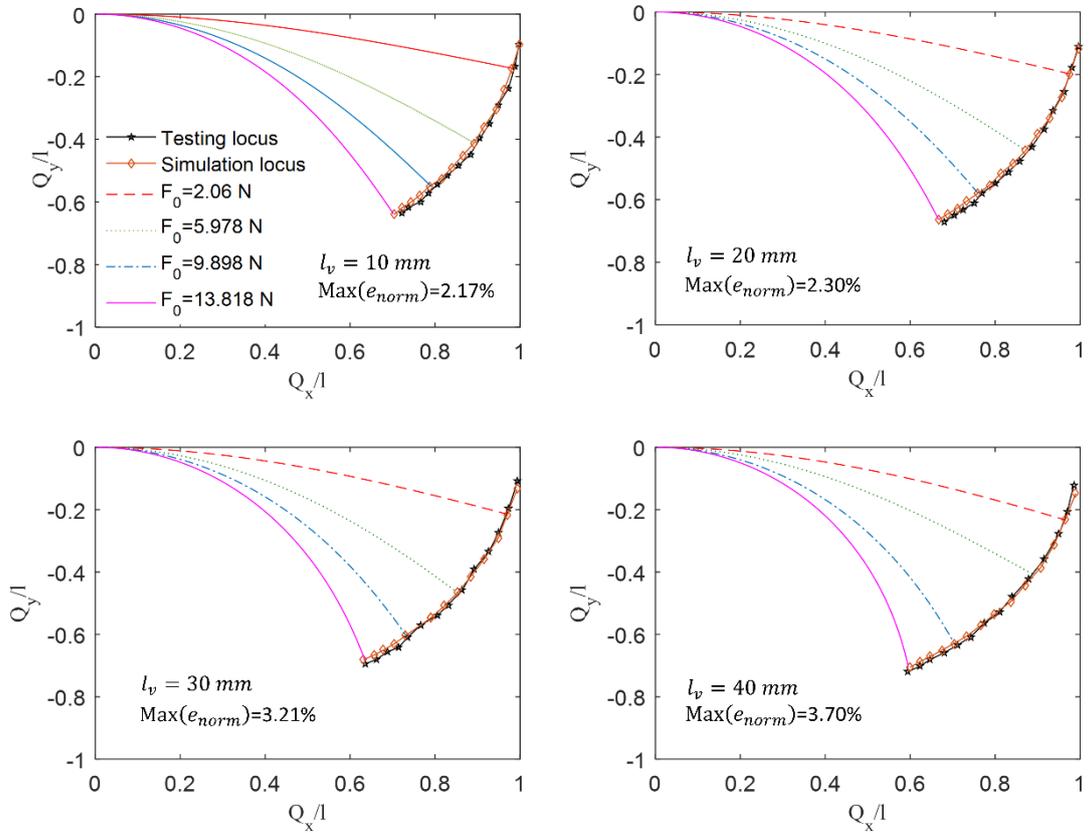

**Fig. 7 Comparison of the tip locus of the simulation and testing results. (Uniform cantilever beam subjected to a combined force and moment load)**

In case 1, angle ∅ can be regulated through adjusting the inclination angle of the cantilever beam, as shown in **Fig. 4(a)**. Here, four different angles ∅ are investigated, and the tip force ranges from 1.078 N to 13.818 N. The force load consists of a set of weights. The comparison between the simulation results and testing results is presented in **Fig. 6**. The maximum transverse deflection can reach about 0.75 when angle ∅ is $-\frac{5\pi}{6}$, and the maximum error is 3.97% when angle ∅ is $-\frac{2\pi}{3}$. While angle ∅ is $-\frac{\pi}{3}$, the maximum transverse deflection is about 0.45, and the maximum error is about 1.44%.

In case 2, the cantilever beam experiences a combined tip force and moment load. Angle ∅ is fixed to $-\frac{\pi}{2}$, but the ratio of the force to the moment can be changed through adjusting the length of the lever arm as shown in **Fig. 4(b)**. The pulley fixed to the tip of the beam is considered as an extra weight, $F_e$. In this case, the lever arm ranges from 10 mm to 40 mm. When the length of the lever arm is 10 mm, the maximum transverse deflection is 0.64, and the maximum error is 2.17%, as shown in **Fig. 7**. While the length of the lever arm is increased to 40 mm, the maximum transverse deflection is augmented to 0.71, and the maximum error can reach 3.70%. In addition, a set of deflection curves when the non-uniform cantilever beam experiences different combined tip force and moment loads are also plotted in **Fig. 7**. As the directions of the tip force and tip moment are the same, no inflection point occurs in the deflection curve.

Given the results shown in **Figs. 6** and **7**, we can conclude that the presented optimization algorithm-based approach can precisely predict the large deformation of a uniform cantilever beam when it is subjected to a combined tip load. Even the inclination angle of the force and ratio of force to moment suffer from large variations, the maximum error between the estimated values and the tested results is limited to 3.97%.

**3.2 Deflection of non-uniform beam**

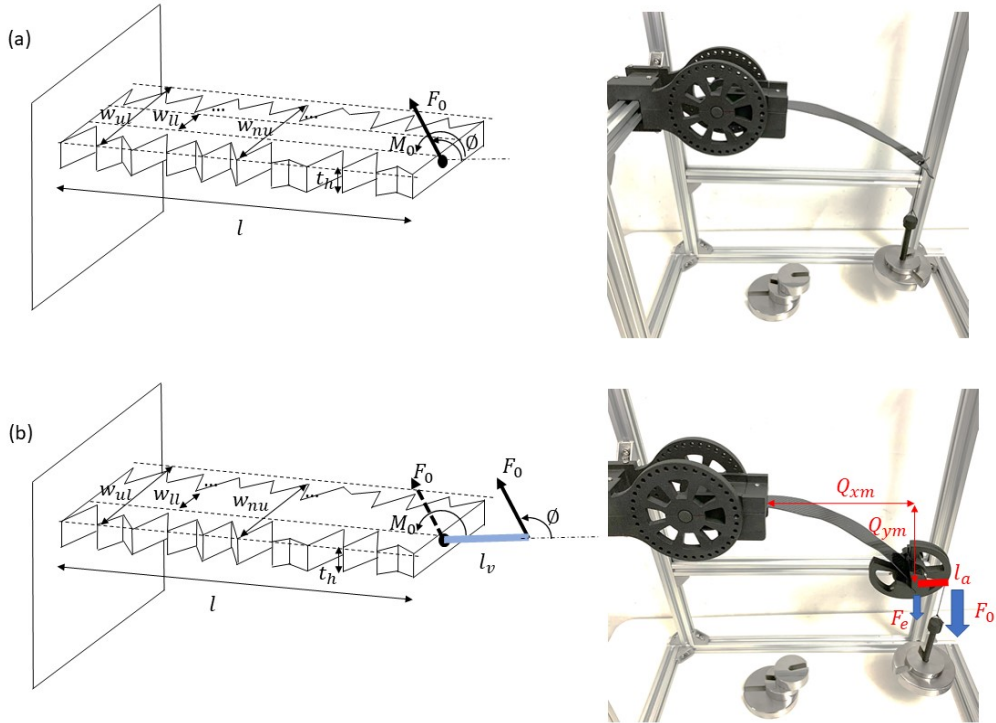

**Fig. 8 Deflection measurements of a non-uniform cantilever beam. (a) Subjected to a pure force load, (b) Subjected to a combined force and moment load.**

The presented method is also validated by characterizing the deflection of a non-uniform cantilever beam. During all testing, plastic deformation should be avoided. After removing the tip loads, the non-uniform cantilever beam can restore to its initial shape, and no residual deformation is induced. Here, the non-uniform beam is also decomposed into *nl* units, and the relationship between the moment and the curvature as well as the coordinates of each unit point are the same as shown in Equations (6)-(10). Unlike in the uniform beam, $I_{l_i}$ in the non-uniform beam is variable rather than constant. As an example, the non-uniform beam is separated into 10 segments, as shown in **Fig. 8**, and the coordinates of the 11 segment points are randomly generated, listed in Table I [57].

$$w_{nu}(i) = w_{ll} + rand * (w_{ul} - w_{ll}), \quad (i = 0,1,2,\cdots,10) \qquad (17)$$

where $w_{ll}$ and $w_{ul}$ denote the lower and upper limits of the width, respectively. In this subsection, also two cases including subjected to a pure force load and subjected to a combined tip force and moment load are studied. In case 1, four different angles $\emptyset$ including $-\frac{\pi}{3}$, $-\frac{\pi}{2}$, $-\frac{2\pi}{3}$, and $-\frac{5\pi}{6}$, are investigated, and the tip force ranges from 1.078 N to 13.818 N. The simulation and testing results for different loads are plotted in **Fig. 9**. When angle $\emptyset$ is $-\frac{\pi}{3}$, the maximum transverse deflection is 0.443, and the maximum error is 1.73%. Then the maximum transverse deflection is increased to 0.746 when angle is changed to $-\frac{5\pi}{6}$. The maximum error is 3.63%.

When the non-uniform cantilever beam suffers from a combined force and moment load, the locus of the beam tip is plotted in **Fig. 10**. The ratio of the force to the moment is adjusted by changing the length of the lever arm, as shown in **Fig. 8(b)**. In this case, angle $\emptyset$ is fixed to $-\frac{\pi}{2}$, and four different lever arms are studied. The pulley fixed to the tip of the beam is considered as an extra weight, $F_e$. When the length of the lever arm is set to 40 mm, the maximum transverse deflection is 0.696, and the maximum error is 4.35%. In addition, a set of deflection curves when the non-uniform cantilever beam experiences different loads are also plotted in **Fig. 10**.

According to the results plotted in **Figs. 9** and **10**, we can conclude that the proposed optimization algorithm-based approach can characterize the large deflection of a non-uniform cantilever beam when subjected to tip loads with high accuracy. Despite changing the inclination angle of the force and the ratio of force to moment, the maximum error is limited to 4.35%.

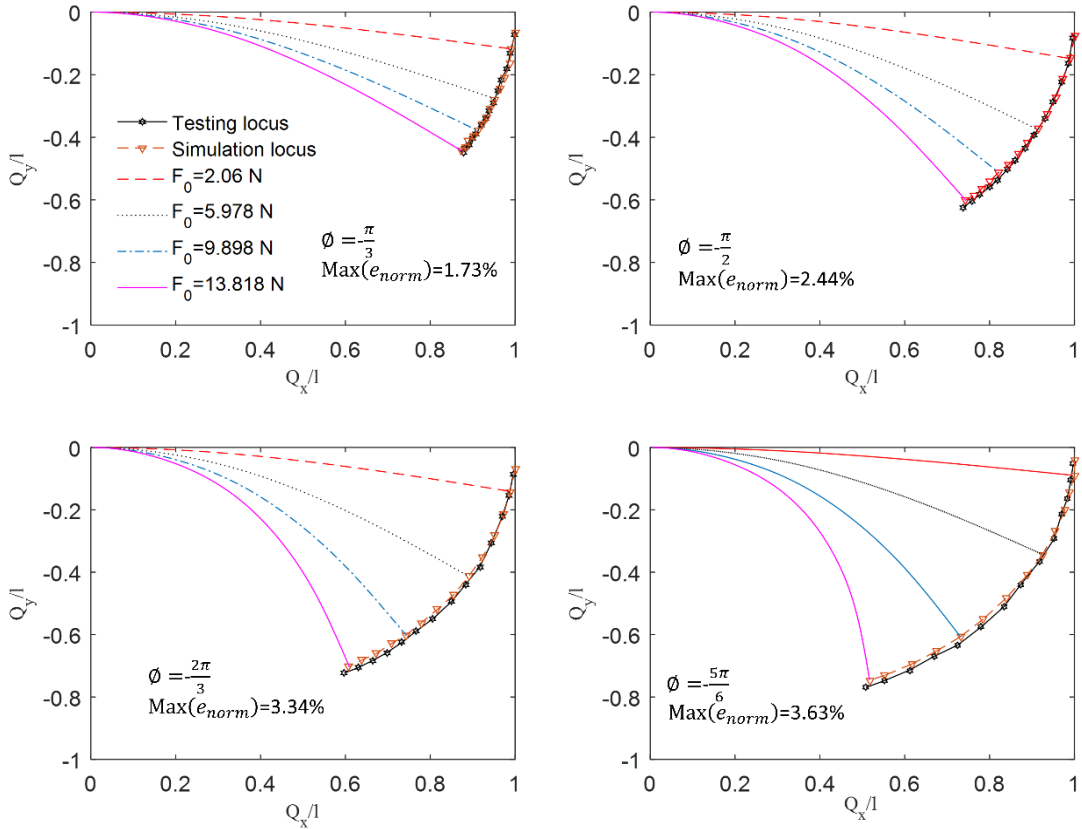

**Fig. 9 Comparison of the tip locus of the simulation and testing results. (Non-uniform cantilever beam subjected to a pure force load)**

## 4. Discussions

Note that as an example, in this paper, we implement a PSO algorithm to find the locus of the tip when the cantilever beam is subjected to different tip loads. Particles in the swarm of the PSO algorithm can gradually approach the solution through updating their own positions and velocities based on the "intelligence" -collective behaviour. In addition to PSO algorithm, the optimization algorithm can be other types of algorithm with the same function such as genetic optimization, ant colony optimization, and neural network algorithms.

In the unit model, equations (6)-(10), only the bending deformation is considered here. However, in some cases, when the cantilever beam is made of stretchable materials, the elongation, shear deformation, or compression deformation in longitudinal direction cannot be ignored [14, 19, 58-59]. To handle that, we need to modify the unit model, equations (6)-(10), to consider the deformation along the neutral plane.

Then we still can employ the presented optimization algorithm (OABA) to find the locus of the tip. This part will be further studied in the future.

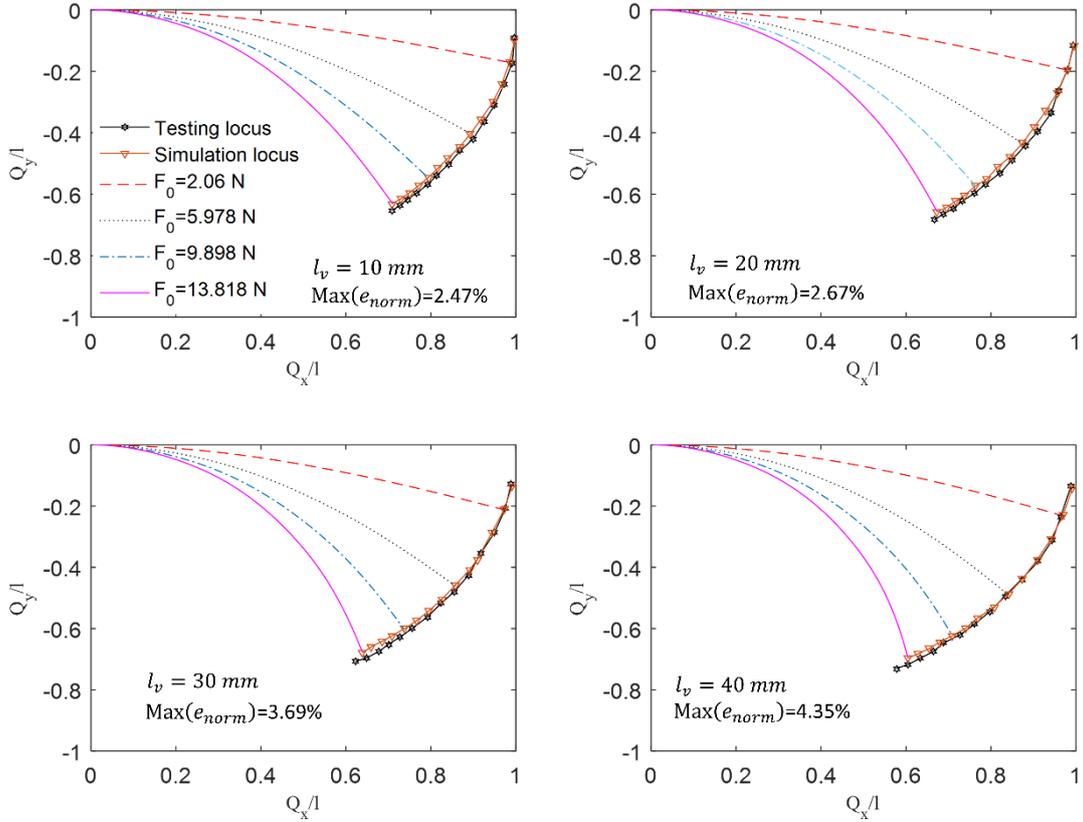

**Fig. 10 Comparison of the tip locus of the simulation and testing results. (Non-uniform cantilever beam subjected to a combined force and moment load)**

The cantilever beam studied in this work is initially straight and without curvature. However, in some cases, the cantilever beam is with initial curvature, as studied in [60-61]. The curvature-moment relationship in a unit, equation (6), should be updated to consider the initial curvature. The feasibility of applying the presented OABA to predict the deflection of a cantilever beam with initial curvature will be verified.

In addition, this presented method is used to characterize the kinematic deformation of the cantilever beam here. In future work, we will check whether this method can be employed to predict the dynamic motion of a cantilever beam [57, 62-65]. The inertia force and damping force will be considered in the unit model.

## 5. Conclusions

In this paper, we proposed a novel method, called optimization algorithm-based approach (OABA), to predict/characterize the large deformation of a cantilever beam when subjecting tip loads. Unlike the other existing approaches, this approach utilized an optimization algorithm (PSO algorithm) to find the locus of the beam tip. When the stopping criteria is fulfilled, the iteration process will be terminated, and then coordinates of the tip are obtained. Substituting the coordinates of the tip into the unit model, the deflection curve of the cantilever beam can be derived. To validate the presented method, we developed a platform to test the deflection of a uniform and a non-uniform cantilever beams when they experience a pure force load and a combined force and moment load, respectively. Comparing the simulated and testing results, we can find that the presented method can predict the deflections of the uniform and the non-uniform cantilever beams with high accuracy. In particular, the maximum error is limited to 3.97% for the uniform cantilever beam when the maximum transverse deflection reaches 0.75. While for the non-uniform beam, the maximum error is 4.35% when the maximum transverse deflection reaches 0.746. Given that, the OABA method can be regarded as a universal method for modelling the large deflection of uniform and non-uniform beams.

## Acknowledgements

This work is supported by Research Grants Council (Project No. CUHK14210019) and the Innovation and Technology Commission (Project No. ITS/367/18) of Hong Kong Special Administrative Region, China and in part by Hong Kong Centre for Logistics Robotics of InnoHK.

# Appendix A

Codes in MATLAB for implementing the PSO-based approach

*(1) Main.m*
```
clear;
clc;
clear global leng thic wid E Q_x Q_y F0 M0 fai theta0 numbs inertia errori xt Pbest;

%%%%% values of the parameters of the cantilever beam %%%%%%%%
global leng thic wid E Q_x Q_y F0 M0 fai theta0 numbs inertia Pbest
leng=180*10^(-3);                       %length of the cantilever beam
thic=1.15*10^(-3);                      %thickness of the cantilever beam
E=45.36*10^9;                           %Young's modulus of the cantilever beam
numbs=200;                              %nums:number of the unit segments

mode=1;                                 %model=1 uniform cantilever beam,
                                        %model=2 non-uniform cantilever beam
if mode==1;
    for i=1:numbs
        wid(i)=25*10^(-3);              %uniform cantilever beam
    end
else
    n_sigm=10;                          %number of the segment
    wid_n=[22.6,26.9,30.0,22.8,22.8,26.8,28.1,20.8,23.7,28.1,29.6]*10^(-3);
    for i=1:numbs
        nl=floor(i/(numbs/n_sigm));
        nd=rem(i,(numbs/n_sigm));
        if nd>0
            wid(i)=wid_n(nl+1)+nd/(numbs/n_sigm)*(wid_n(nl+2)-wid_n(nl+1
        else
            wid(i)=wid_n(nl+1);
        end
    end
end
inertia=1/12.*wid*thic.^3;              %moment of inertia
                                        %(Q_x Q_y)the coordinates of the tip
F_i=20;
F0=F_i*9.8*50*10^(-3)+9.8*10*10^(-3);   %F0: tip force
lever_a=0;
M0=-F0*lever_a*10*10^(-3);              %M0: tip moment
fai_i=5;
fai=fai_i*(-pi/6);                      %fai: the direction of the tip Force
                                        %theta0: the tip slope angle

%%%%% solving processes %%%%%%%%%%%
global errori xt
xL=[-leng leng;-leng leng; -pi pi];     %the upper and lower limits of the parameters
                                        %Q_x, Q_y, theta0 (output of PSO)
N=100;                                  %number of particles in PSO
c1=0.2;                                 %constant parameters in PSO
c2=c1;                                  %constant parameters in PSO
wmax=0.8;                               %the maximum inertia weight in PSO
wmin=0.6;                               %the minimum inertia weight in PSO
max_step=50;                            %the maximum iteration number
vd=3;                                   %dimension of the input in PSO
[xm,fv] = PSO(@inversemodel,N,c1,c2,wmax,wmin,max_step,vd,xL);
```

*(2) PSO.m*
…………………..

*(3) unit_model.m*
………………….. (All codes will be provided after this manuscript has been accepted for publication)